\documentclass[prd, onecolumn]{revtex4}

\usepackage{graphicx}
\usepackage{dcolumn}
\usepackage{bm}
\usepackage{wrapfig}
\usepackage{epsfig}
\usepackage{breqn}
\usepackage{amssymb}
\usepackage{subfigure}
\usepackage{float}
\usepackage{pdfpages}

\begin{document}


\title{The Wavefunction of Continuous-Time Recurrent Neural Networks}

\author{Ikjyot Singh Kohli}
	\email{isk@mathstat.yorku.ca}
\affiliation{York University - Department of Mathematics and Statistics}
\author{Michael C. Haslam}
	\email{mchaslam@mathstat.yorku.ca}
\affiliation{York University -  Department of Mathematics and Statistics}

\date{February 12, 2020}

\begin{abstract}
In this paper, we explore the possibility of deriving a quantum wavefunction for continuous-time recurrent neural network (CTRNN). We did this by first starting with a two-dimensional dynamical system that describes the classical dynamics of a continuous-time recurrent neural network, and then deriving a Hamiltonian. After this, we quantized this Hamiltonian on a Hilbert space $\mathbb{H} = L^2(\mathbb{R})$ using Weyl quantization. We then solved the Schrodinger equation which gave us the wavefunction in terms of Kummer's confluent hypergeometric function corresponding to the neural network structure. Upon applying spatial boundary conditions at infinity, we were able to derive conditions/restrictions on the weights and hyperparameters of the neural network, which could potentially give insights on the the nature of finding optimal weights of said neural networks.
\end{abstract}
\maketitle 

\section{Introduction}
Continuous-time recurrent neural networks (CTRNNs) have been increasingly employed in the field of machine learning and deep learning to understand data that evolves in continuous time. This is largely due to the fact that CTRNNs exhibit rich temporal dynamics and are suitable for such tasks. Indeed, Harvey, Husbands, and Cliff \cite{inproceedingsHarvey} used CTRNNs to successfully model robotic vision tasks in which agents were able to robustly perform simple visually guided tasks. Further, Beer \cite{BEER1997257} described in detail how CTRNNs could be used to accurately model agents that exhibit adaptive and cognitive behaviour.

Although artificial neural networks in general are found to model key cognitive aspects of brain function, such as learning and classification, their black-box nature has led many to gain deeper insights as to how and why artificial neural networks especially in the context of deep learning work so well in problems of learning and classification. Olah \cite{olah1} has described a heuristic approach using the Manifold Hypothesis to explain the so-called ``inner workings'' of a neural network. Further, Lin, Tegmark and Rolnick \cite{lintegmark} used various properties from statistical and mathematical physics to describe how a deep neural network may actually work.

Motivated by the aforementioned work and questions relating to how a neural network may \emph{actually} work, we analyze a specific class of artificial neural networks, namely continuous-time recurrent neural networks (CTRNNs). We specifically consider the case of two nodes in which the dynamical system describing the evolution of these nodes is planar, that is, in $\mathbb{R}^2$. We further consider the case where the activation function is the rectified linear unit function also known as the ReLU function. An important point to note is that while it is fairly standard in the literature to use sigmoid activation functions in the contexts of time-dependent recurrent neural networks as in \cite{doi:10.1177/105971239500300405}, other activation functions have been used in a variety of different problems. For example, Ianigro and Bown \cite{ianigro2018exploring} used an inverse hyperbolic tangent activation function in CTRNNs to analyze the structure of audio waveforms. 

The paper is organized as follows. We first establish the conditions under which the CTRNN equations can be written in Hamiltonian form. After a description of this structure, we then proceed with a discussion on how such a system can be quantized and solve an analogue of the Schrodinger equation, thereby establishing the wavefunction of quantum mechanics as the main low-level mechanism responsible for the observed dynamics of such CTRNNs. We feel that this could provide great insights into the nature of the weights found during the neural network training procedure. Typically, finding optimal weights amounts to finding a global minimum of a loss function which given the nature of these loss functions, is very difficult to do. One typically has to rely on iterative numerical methods such as gradient descent to minimize such loss functions, but, in most cases, one has to settle for local minima or saddle points. Some of these issues are discussed in \cite{baba1989new}, \cite{shang1996global}, \cite{dauphin2014identifying}, and \cite{pascanu2014saddle}.

\section{Deriving A Hamiltonian}
It is well-known that a continuous-time recurrent neural network (CTRNN) can be described by a generally non-linear system of
ordinary differential equations \cite{doi:10.1177/105971239500300405}:
\begin{equation}
\label{eq:dynsys1}
\tau_i \dot{y}_{i} = -y_i + \sum_{j=1}^{n} w_{ji} \sigma(y_j - \theta_j) + I_i(t),
\end{equation}
where $i$ is the neuron index, $y_i$ is the action potential, $\tau_i$ is the time constant of the postsynaptic node, $w_{ji}$ are the 
connection weights between nodes, $\sigma(x)$ is the nonlinear activation function, $\theta_j$ is a bias term, and $I_i(t)$ denote the input
to a particular node. Note that in the work that follows, we consider an autonomous system of ordinary differential equations thereby considering the case where $I_i(t) = I_i$ which are time-independent constants.

We consider the case of two neurons/nodes, in which Eqs. (\ref{eq:dynsys1}) reduces to a planar dynamical system:
\begin{eqnarray}
\label{eq:dyn1}
\dot{y}_{1} &=& \frac{1}{\tau_1} \left[-y_1 + w_{11} f\left(y_1 + \theta_1\right) + w_{21} f\left(y_2 + \theta_2\right) + I_1 \right], \\
\label{eq:dyn2}
\dot{y}_{2} &=& \frac{1}{\tau_2} \left[-y_2 + w_{22} f \left(y_2 + \theta_2 \right) + w_{12} f \left(y_1 + \theta_1 \right) + I_2 \right],
\end{eqnarray}
where $f(z)$ is an activation function. For this paper, we will take $f(z)$ to be the rectified linear unit function (also known as ReLU) which is defined as
\begin{equation}
\label{eq:relu}
f(z) = z^{+} = \max(0,z).
\end{equation}
This means that Eqs. (\ref{eq:dyn1})-(\ref{eq:dyn2}) take the form:
\begin{eqnarray}
\label{eq:dyn11}
\dot{y}_{1} &=& \frac{1}{\tau_1} \left[-y_1 + w_{11}\left(y_1 + \theta_1\right) + w_{21} \left(y_2 + \theta_2\right) + I_1\right], \\
\label{eq:dyn22}
\dot{y}_{2} &=& \frac{1}{\tau_2} \left[-y_2 + w_{22}\left(y_2 + \theta_2 \right) + w_{12} \left(y_1 + \theta_1\right) + I_2 \right],
\end{eqnarray}
if $y_1 + \theta_1 \geq 0$ and $y_2 + \theta_2 \geq 0$,
or
\begin{eqnarray}
\label{eq:dyn111}
\dot{y}_{1} &=& \frac{1}{\tau_1} \left[-y_1 + I_1\right], \\
\label{eq:dyn222}
\dot{y}_{2} &=& \frac{1}{\tau_2} \left[-y_2 + I_2\right],
\end{eqnarray}
if $y_1 + \theta_1 < 0$ and $y_2 + \theta_2 < 0$.
Clearly, the dynamics in the case of Eqs. (\ref{eq:dyn111})-(\ref{eq:dyn222}) are very simple as the equations are decoupled. In fact, they imply that:
\begin{eqnarray}
y_1(t) &=& I_1 + C_1\exp\left(-\frac{t}{\tau_1}\right), C_1 \in \mathbb{R}, \\
y_2(t) &=& I_2 + C_2 \exp\left(-\frac{t}{\tau_2}\right), C_2 \in \mathbb{R}.
\end{eqnarray}

We are now interested to see if Eqs. (\ref{eq:dyn11})-(\ref{eq:dyn22}) can be written as a Hamiltonian system. Recall, for a smooth function $H: \mathbb{R}^2 \to \mathbb{R}$, a Hamiltonian system is of the form
\begin{equation}
\dot{y}_{1} = \frac{\partial H}{\partial y_2}(y_1,y_2) \quad \dot{y}_{2} = -\frac{\partial H}{\partial y_1}(y_1, y_2),
\end{equation}
where the function $H(y_1, y_2)$ is referred to as the Hamiltonian function.

Let $f_1$ and $f_2$ denote the right-hand-sides of Eqs. (\ref{eq:dyn11})-(\ref{eq:dyn22})  respectively. Then, Eqs. (\ref{eq:dyn11})-(\ref{eq:dyn22})  constitute a Hamiltonian system if
\begin{equation}
\frac{\partial f_1}{\partial y_1} = - \frac{\partial f_2}{\partial y_2}.
\end{equation}
This implies that
\begin{equation}
\label{eq:ham1}
\frac{-1 + w_{11}}{\tau_1} = \frac{1-w_{22}}{\tau_2}.
\end{equation}
This is true in three cases. Keeping mind that $\tau_{1}, \tau_{2} > 0$, we have the first possibility that $w_{22} = 1$ and $w_{11} = 1$. The second and third possibilities occur when:
\begin{equation}
\tau_{1} = \frac{\tau_2 - \tau_2 w_{11}}{-1 + w_{22}}, \quad w_{22} < 1, w_{11} > 1,
\end{equation}
\begin{equation}
\tau_{1} = \frac{\tau_2 - \tau_2 w_{11}}{-1 + w_{22}}, \quad w_{22} > 1, w_{11} < 1.
\end{equation}
It is of interest to note that $w_{12}$ and $w_{21}$  are not bound by these Hamiltonian conditions. 

We can derive a Hamiltonian for the general case as follows. We integrate the right hand side of Eq. (\ref{eq:dyn11}) with respect to $y_2$ and obtain:
\begin{equation}
\label{eq:H1}
H(y_1,y_2) =  \frac{w_{21}}{2 \tau_1}y_2^2 + \frac{y_1y_2}{\tau_1}(w_{11}-1)
+\frac{y_2}{\tau_1} \left(w_{11}\theta_1 + w_{21}\theta_2+I_1\right)+ g(y_1),
\end{equation}
where $g(y_1)$ is an arbitrary function of $y_1$ which we derive now.
Differentiating this equation and equating it with the negative right-hand-side of Eq. (\ref{eq:dyn222}), one obtains that
\begin{equation}
\label{eq:constr1}
g(y_1) =  -\frac{w_{12}}{2 \tau_2}y_1^2
-\frac{y_1}{\tau_2} \left(w_{22}\theta_2+w_{12}\theta_1+I_2\right) + const.
\end{equation}
Substituting Eq. (\ref{eq:constr1}) into Eq. (\ref{eq:H1}), we see that the Hamiltonian of the system is given by (neglecting the constant of integration):
\begin{equation}
\label{eq:H2}
H(y_1, y_2) =  -C_1 y_1 - C_2 y_1^2 + D_1 y_2 + D_2 y_2^2+E_2y_1y_2,
\end{equation}
where, for convenience, we define constants
\begin{equation}
\label{eq:const1}
C_1 = \frac{1}{\tau_2}(w_{12}\theta_1+w_{22}\theta_2+I_2),\quad 
D_1 = \frac{1}{\tau_1}(w_{21}\theta_2+w_{11}\theta_1+I_1),\quad
C_2 = \frac{w_{12}}{2\tau_2},\quad
D_2 = \frac{w_{21}}{2\tau_1},
\end{equation}
and 
\begin{equation}
\label{eq:const2}
E_2 = \frac{1}{\tau_1}(w_{11}-1) = -\frac{1}{\tau_2}(w_{22}-1).
\end{equation}
A contour plot of the Hamiltonian given by Eq. (\ref{eq:H2}) is shown in Fig. \ref{Fig:fig1} below.
\begin{figure}[h]
  \centering
    \includegraphics{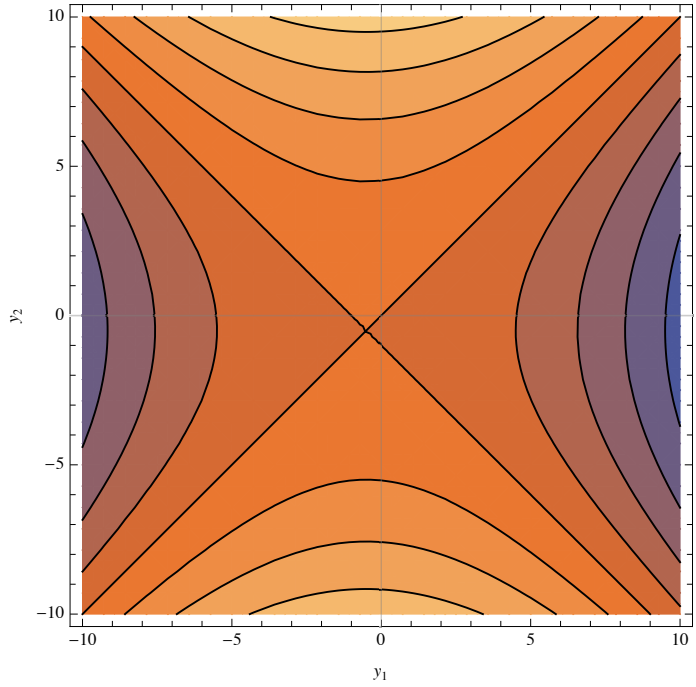}
    \caption{A Contour plot of the Hamiltonian described by Eq. (\ref{eq:H2}). One sees that the system exhibits similar behaviour to that of a classical nonlinear pendulum.}
     \label{Fig:fig1}
\end{figure}

We know from quantum mechanics that the behaviour of classical systems is determined by underlying quantum behaviour. The classical behaviour of the above neural networks can therefore ideally be understood by the underlying wavefunction, which we attempt to now derive. We now turn to the question of quantizing Eq. (\ref{eq:H2}). Let us consider a Hilbert space $\mathbb{H} = L^2(\mathbb{R})$. We assume that the classical observables $y_{1,2}$ can be taken to self-adjoint operators $\hat{y}_{1,2}$, such that
\begin{equation}
\label{eq:operators}
\hat{y}_{2} = y_2, \quad \hat{y}_{1} = -i \frac{\partial}{\partial y_2},
\end{equation}
where $i = \sqrt{-1}$. 


We apply Weyl quantization to Eq. (\ref{eq:H2}) to obtain a symmetric quantized Hamiltonian:
\begin{equation}
\label{eq:H3}
\hat{H} = i C_1 \frac{\partial}{\partial y_2}
- C_2 \frac{\partial^2}{\partial y_2^2} + D_1 y_2 + D_2 y_2^2 + E_2 \left[-\frac{i}{2} -i \frac{y_2}{2} \frac{\partial}{\partial y_2}. \right]
\end{equation}

\section{Deriving The Wavefunction}
We see from Eq. (\ref{eq:H3}) that the Hamiltonian operator does not have an explicit time dependence. We can therefore derive the wavefunction using the time-independent Schrodinger eigenvalue problem: $\hat{H} |\Psi\rangle = \lambda | \Psi \rangle$, where $\lambda$ is the eigenvalue corresponding to the Hamiltonian operator $\hat{H}$. In physical problems, $\lambda$ is the energy of the system.

The time-independent Schrodinger equation then implies the following eigenvalue problem:
\begin{equation}
C_2\frac{d^2 \psi(y_2)}{dy_2^2} +  \left(C_1-E_2y_2\right) \frac{d \psi(y_2)}{dy_2} +
(D_1 y_2 - D_2 y_2^2-E_2)\psi(y_2) = \lambda\psi(y_2),
\end{equation}
where the eigenfunctions $\psi$ must obey conditions $\psi\rightarrow 0$ as $|y_2|\rightarrow \infty$, which ensures the eigenfunctions are normalizable. In quantum mechanics, this condition implies that a particle with finite energy cannot move to arbitrarily large distances. Formal solutions of the differential equation have the form
\begin{equation}
\psi(y_2) = \Psi(y_2){\mathrm{e}}^{-y_2(\xi_1 y_2+\xi_2)}
\end{equation}
with
\begin{equation}
\label{eq:xi1}
\xi_1 = 4C_2D_2+E_2^2-E_2\sqrt{4C_2D_2+E_2^2}
\end{equation}
and
\begin{equation}
\label{eq:xi2}
\xi_2 = -C_1E_2-4C_2D_1+2\sqrt{4C_2D_2+E_2^2},
\end{equation}

\begin{eqnarray}
\label{eq:specialfunc1}
\Psi(x) = A_1 M\!\left(\alpha,\frac{1}{2},\frac{\left(-(4C_2D_2+E_2^2)x+C_1E_2+2C_2D_1\right)^2}
{(4C_2D_2+E_2^2)^{3/2}}
\right)\\+A_2 \left[(4C_2D_2+E_2^2)x-2D_1C_2-C_1E_2\right]
M\!\left(\beta,\frac{3}{2},\frac{\left(-(4C_2D_2+E_2^2)x+C_1E_2+2C_2D_1\right)^2}
{(4C_2D_2+E_2^2)^{3/2}}
\right).
\end{eqnarray}
We have further denoted by $A_1$ and $A_2$ arbitrary constants, and parameters given by
\begin{equation}
\alpha = \frac{1}{4} + \frac{2(4C_2D_2+E_2^2)\lambda+2D_2C_1^2-2C_1D_1E_2-2C_2D_1^2+4C_2D_2E_2+E_2^3
}{4(4C_2D_2+E_2^2)^{3/2}}
\end{equation}
and
\begin{equation}
\beta = \alpha+\frac{1}{2}.
\end{equation}

As per Eq. (\ref{eq:specialfunc1}), we see that the wavefunction solution depends on the special function $M(.)$, which denotes Kummer's function. As is well-known, Kummer's function is defined by the series expansion
\begin{equation}
M(a,b,z) = ~_1F_1\!\left(a;b;z\right) = \sum_{n=0}^{\infty}\frac{(a)_nz^n}{n!(b)_n}
\end{equation}
where $(a)_n$ denotes the Pochhammer symbol (the rising factorial). Interestingly, this infinite series expansion terminates (thus forming a hypergeometric polynomial) when $a$ is a non-positive integer. It is precisely this condition that is required to ensure that solutions of the differential equation above obey the conditions as  $|y_2|\rightarrow \infty$. Thus, setting $\alpha=-m$ and $\beta=-m$ for $m=0,1,2,\dots$ produces a discrete (but infinite) set of eigenvalues and eigenfunctions.

In particular, the even and odd eigenfunctions are
\begin{equation}
\psi_{2m}(y_2) = H_{2m}\!\left(\frac{-(4C_2D_2+E_2^2)y_2+C_1E_2+2C_2D_1}{\sqrt{2C_2}(4C_2D_2+E_2^2)^{3/4}}\right){\mathrm{e}}^{-y_2(\xi_1 y_2+\xi_2)}
\end{equation}
and
\begin{equation}
\psi_{2m+1}(y_2) = H_{2m+1}\!\left(\frac{-(4C_2D_2+E_2^2)y_2+C_1E_2+2C_2D_1}{\sqrt{2C_2}(4C_2D_2+E_2^2)^{3/4}}\right){\mathrm{e}}^{-y_2(\xi_1 y_2+\xi_2)},
\end{equation}
where the functions $H_{2m}$ and $H_{2m+1}$ denote the Hermite polynomials. The necessary relationships between the hypergeometric functions and the Hermite polynomials are given by
\begin{equation}
H_{2n}(x) = (-1)^n\frac{(2n)!}{n!}~_1F_1\!\left(-n;\frac{1}{2};x^2\right)
\end{equation}
and
\begin{equation}
H_{2n+1}(x) = 2x(-1)^n\frac{(2n+1)!}{n!}~_1F_1\!\left(-n;\frac{3}{2};x^2\right).
\end{equation}

The eigenvalues corresponding to these even and odd eigenfunctions are
\begin{equation}
\lambda_m^{e} = -\frac{1}{2}\frac{(4m+1)(4C_2D_2+E_2^2)^{3/2}+2D_2C_1^2-2C_1D_1E_2-2C_2D_1^2+4C_2D_2E_2+E_2^3}{4C_2D_2+E_2^2}
\quad m=0,1,2,\dots
\end{equation}
and
\begin{equation}
\lambda_m^{o} = -\frac{1}{2}\frac{(4m+3)(4C_2D_2+E_2^2)^{3/2}+2D_2C_1^2-2C_1D_1E_2-2C_2D_1^2+4C_2D_2E_2+E_2^3}{4C_2D_2+E_2^2}
\quad m=0,1,2,\dots.
\end{equation}
We note that for the eigenfunctions to obey the conditions $\psi(y_2)\rightarrow 0$ as $|y_2|\rightarrow \infty$ we must impose a few more conditions on the exponential terms. In particular we require $\mathrm{Real}(\xi_1)>0$ or $\mathrm{Real}(\xi_1)=0$ and $\mathrm{Real}(\xi_2)>0$. These conditions are very interesting as they place restrictions on values of the neural network weights and hyperparameters. For example, this can be seen from the condition $\mathrm{Real}(\xi_1)>0$ . In fact, Eqs.  (\ref{eq:const1})-(\ref{eq:const2}) and (\ref{eq:xi1}) imply that:
\begin{equation}
\label{eq:restr1}
\frac{\left(w_{11}-1\right){}^2}{\tau _1^2}-\frac{\left(w_{11}-1\right) \sqrt{\frac{\left(w_{11}-1\right){}^2}{\tau _1^2}+\frac{w_{12} w_{21}}{\tau _1 \tau _2}}}{\tau _1}+\frac{w_{12} w_{21}}{\tau _1 \tau _2} > 0,
\end{equation}
which is the condition $\mathrm{Real}(\xi_1)>0$. While, the condition $\mathrm{Real}(\xi_1)=0$ is explicitly given by:
\begin{equation}
\label{eq:restr2}
\frac{\left(w_{11}-1\right){}^2}{\tau _1^2}-\frac{\left(w_{11}-1\right) \sqrt{\frac{\left(w_{11}-1\right){}^2}{\tau _1^2}+\frac{w_{12} w_{21}}{\tau _1 \tau _2}}}{\tau _1}+\frac{w_{12} w_{21}}{\tau _1 \tau _2} = 0.
\end{equation}
Further, using Eqs.  (\ref{eq:const1})-(\ref{eq:const2}) and (\ref{eq:xi2}), we see that the condition $\mathrm{Real}(\xi_2)>0$ is explicitly given by:
\begin{equation}
\label{eq:restr3}
-\frac{2 w_{12} \left(I_1+\theta _1 w_{11}+\theta _2 w_{21}\right)}{\tau _1 \tau _2}-\frac{\left(w_{11}-1\right) \left(I_2+\theta _1 w_{12}+\theta _2 w_{22}\right)}{\tau _1 \tau _2}+2 \sqrt{\frac{\frac{\tau _1 w_{12} w_{21}}{\tau _2}+\left(w_{11}-1\right){}^2}{\tau _1^2}} > 0.
\end{equation}

What is very interesting is that the boundary conditions that describe the vanishing of the wavefunction at spatial infinity impose the conditions described in Eqs. (\ref{eq:restr1})-(\ref{eq:restr3}). The reason why this is interesting is because finding optimal weights to train a neural network is one of the most important (and open!) problems in the fields of artificial neural networks and deep learning. As mentioned in the introduction, this is quite difficult to do in most cases due to the potential highly nonlinear and complicated nature of loss functions associated with a given neural network. Often, only iterative numerical algorithms such as gradient descent can be used, which seldom leads to global minima. This is discussed in length in \cite{ludermir2006optimization}, \cite{pavelka2004algorithms}, and more generally in \cite{goodfellow2016deep}. This work shows that the values of these weights could actually be motivated by an underlying quantum structure.




\section{Discussion}
In this paper, we have explored the possibility of deriving a quantum wavefunction for continuous-time recurrent neural network (CTRNN). We did this by first starting with a two-dimensional dynamical system that describes the classical dynamics of a continuous-time recurrent neural network, and then deriving a Hamiltonian. After this, we quantized this Hamiltonian on a Hilbert space $\mathbb{H} = L^2(\mathbb{R})$ using Weyl quantization. We then solved the Schrodinger equation which gave us the wavefunction corresponding to the neural network structure. Upon applying spatial boundary conditions at infinity, we were able to derive conditions/restrictions on the weights of the neural network, which could potentially give insights on the nature of the weights found during the neural network training procedure.

There are some points that warrant deeper discussion. The first point is how does one quantize a CTRNN when the number of neurons is odd? It seems that the Hamiltonian procedure works well when the number of neurons is even subject to some restrictions such as those in Eq. (\ref{eq:ham1}). However, one may have to use a completely different approach to quantization when the number of neurons is odd. A promising approach is described in \cite{floratos2012matrix} where the three-dimensional Lorenz system was quantized using noncommutative phase space coordinates. 

Another point to consider is that we were only able to describe the quantization procedure when a condition as in Eq. (\ref{eq:ham1}) holds. It is a deep curiosity that we were not able to presently resolve as to why this should be the case. One would expect that the existence of such a quantum mechanical structure should always hold independent of any such condition. It could be that this occurs because of quantizing the system with respect to a Hamiltonian / Schrodinger approach. In this case, one may have to make use of approaches of quantizing non-Hamiltonian systems as described in \cite{tarasov2001quantization} and references therein. We plan to explore this approach in future work.

The final point to consider is the physical interpretation of the work presented in this paper. Clearly, in the overwhelming number of cases, quantization is applied and wavefunctions are derived for physical systems such as harmonic oscillators are particles in a box, where some physical parameter like the mass of the particle is always well-defined. In this case, it is not clear whether a continuous-time recurrent neural network corresponds to a real physical system that can be described by such an equivalent physical parameter. The Hamiltonian we derived in this work came from geometric arguments from dynamical systems, but it is an open question on what a physically representative Hamiltonian looks like for such a neural network system. This issue also then leads to the issue of physically meaningful boundary conditions. For example, in the case of a particle in a box problem, one typically applies Dirichlet boundary conditions to obtain the energy eigenvalues and eigenfunctions. 

As far as we can tell, the aforementioned issues raised have either not been raised in the scientific literature or addressed in any conclusive way. We therefore are hopeful that such questions could lead to more research results in these areas that could lead to interesting answers that could provide even deeper insights on the connections between neural networks and physics.

\bibliography{sources}

\end{document}